\documentclass[conference]{IEEEtran}
\usepackage{blindtext, graphicx}

\ifCLASSINFOpdf
\else
\fi

\usepackage{amsmath,amssymb,amsthm}
\usepackage{cite}
\usepackage{algorithm}
\usepackage{algorithmic}
\usepackage{caption}
\usepackage{ltablex}
\usepackage{booktabs}
\usepackage{hyperref}
\usepackage{subcaption}
\usepackage{url}
\usepackage{color}
\usepackage{graphicx}
\usepackage{xcolor}
\let\labelindent\relax
\usepackage{enumitem}

\usepackage{array}
\usepackage{multicol}

\usepackage{caption}
\usepackage{booktabs}
\usepackage{float}
\usepackage{etoolbox}

\usepackage[bottom]{footmisc}

\usepackage{booktabs}

\usepackage{array}
\newcolumntype{L}[1]{>{\raggedright\let\newline\\\arraybackslash\hspace{0pt}}m{#1}}
\newcolumntype{C}[1]{>{\centering\let\newline\\\arraybackslash\hspace{0pt}}m{#1}}
\newcolumntype{R}[1]{>{\raggedleft\let\newline\\\arraybackslash\hspace{0pt}}m{#1}}
\newcommand{\bx}{\ensuremath{\mathbf{x}}}

\newcommand{\bz}{\ensuremath{\mathbf{z}}}

\newcommand{\by}{\ensuremath{\mathbf{y}}}
\newcommand{\ba}{\ensuremath{\mathbf{a}}}
\newcommand{\tildt}{\ensuremath{\tilde{t}}}

\DeclareMathOperator*{\argmin}{arg\,min}

\newtheorem{definition}{Definition}

\definecolor{dgreen}{RGB}{63, 175, 115}

\begin{document}
\sloppy
\title{Learning Rich Geographical Representations: Predicting Colorectal Cancer Survival in the State of Iowa}


\author{\IEEEauthorblockN{Michael T. Lash\IEEEauthorrefmark{1}, Yuqi Sun\IEEEauthorrefmark{1}, Xun Zhou\IEEEauthorrefmark{2}, Charles F.~Lynch\IEEEauthorrefmark{3}, and W.~Nick Street\IEEEauthorrefmark{2}}
	\IEEEauthorblockA{\IEEEauthorrefmark{1}Department of Computer Science, \IEEEauthorrefmark{2}Department of Management Sciences,
	\IEEEauthorrefmark{3}Department of Epidemiology\\
		University of Iowa\\
		Iowa City, Iowa 52242\\
		\{michael-lash, yuqi-sun, xun-zhou, charles-lynch, nick-street\}@uiowa.edu}
}

\maketitle

\begin{abstract}
Neural networks are capable of learning rich, nonlinear feature representations shown to be beneficial in many predictive tasks. In this work, we use these models to explore the use of geographical features in predicting colorectal cancer survival curves for patients in the state of Iowa, spanning the years 1989 to 2012. Specifically, we compare model performance using a newly defined metric -- \textit{area between the curves} (ABC) -- to assess (a) whether survival curves can be reasonably predicted for colorectal cancer patients in the state of Iowa, (b) whether geographical features improve predictive performance, and (c) whether a simple binary representation or richer, spectral clustering-based representation perform better. Our findings suggest that survival curves can be reasonably estimated on average, with predictive performance deviating at the five-year survival mark. We also find that geographical features improve predictive performance, and that the best performance is obtained using richer, spectral analysis-elicited features.
\end{abstract}



\section{Introduction}

The rise of machine learning  and corresponding advent of various deep learning methodologies in recent years hold great promise as such methods are capable of learning rich, non-linear feature representations. Such representations have been shown to be beneficial in a variety of domains, including medicine and public health. This work is concerned with methodology applied to such areas. More specifically, our focus is on exploring different representations of geographical features that can be used to predict colorectal cancer survival curves for patients in the state of Iowa.

\begin{figure}[h]
    \centering
    \includegraphics[scale=.08]{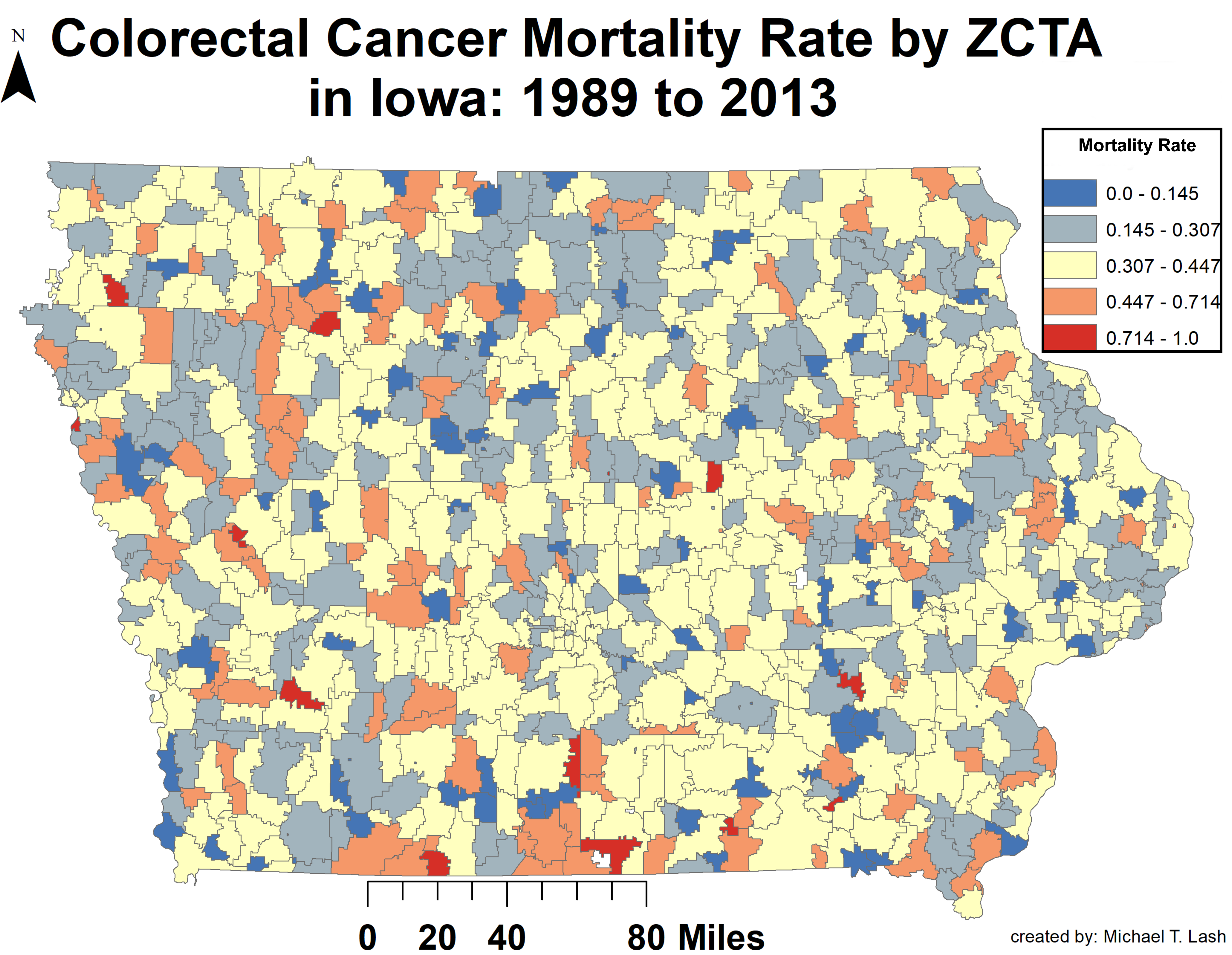}
    \caption{Colorectal cancer mortality rate by ZCTA in the state of Iowa for the years 1989 to 2012.}
    \label{fig:mortrate}
\end{figure}

To elaborate on such a problem, consider Figure \ref{fig:mortrate}, which shows colorectal cancer mortality rates, spanning the years 1989 to 2013, by zipcode tabulation area (ZCTA), for the state of Iowa.
First, we wish to point out that many zipcodes have mortality rates at or above 30\%, which shows the importance of accurately assessing the survival outlook of patients at the time of diagnosis, which may better inform treatment decisions \cite{Zhang2015DM}. Secondly, Figure \ref{fig:mortrate} shows that different geographic locations experience different mortality rates. In other words, location appears to have a bearing on survival outlook.

Survival outlook-disparity by location is, unfortunately, not unexpected. Geographic location has been shown to have an effect on health care access, thereby affecting colorectal cancer survival outlook \cite{wan2013spatial}. Moreover, environmental factors found to increase the likelihood of developing colorectal cancer tend to be spatially grouped (e.g., houses built when lead-based paint was the norm); incorporation of such factors in predictive models have been shown to provide performance improvements 

Therefore, because of the spatially heterogeneous nature of the colorectal cancer mortality manifestation, a major challenge in accurately predicting colorectal cancer patient survival curves is to construct models that are spatially sensitive to patient locale: key factors affecting survival in large cities may be very different from those in rural areas. This work, therefore, explores two different ways of representing geography -- termed \textit{simple binary representation} (SBR) and \textit{rich representation -- spectral analysis} (RR-SA) -- for use as features in constructing neural network-based predictive models.



The contributions of this work are enumerated as follows:
\begin{enumerate}
    \item We investigate whether colorectal cancer patient survival curves can be reasonably predicted for patients in the state of Iowa.
    \item We examine whether geographical features improve the accuracy of survival curve predictions over models trained without the use of geographic features.
    \item We explore a rich representation of geographical features through spectral analysis (RR-SA) of the underlying adjacency graph of the ZCTAs to address the spatial heterogeneity challenge. 
    \item We determine whether the simple binary representation (SBR) or richer, spectral analysis representation (RR-SA) leads to more accurate survival curve predictions.
    \item We propose a new metric -- \textit{area between curves} (ABC) -- to assess the quality of survival curve predictions.
\end{enumerate}

The remainder of this work proceeds with an outline of our methods of representing geographical features and corresponding neural network architecture associated with each such representation (Section II). Next, we discuss our dataset, which contains 46000 Iowan patients diagnosed with colorectal cancer between the years 1989 and 2012, followed by our experiments and results (Section III). Finally, we discuss related work (Section IV) prior to concluding the paper (Section V).

\section{Learning Geographical Representations for Survival Curve Prediction}

In this section we discuss our methodology, where we begin by outlining some preliminary notation and problem facets, followed by a discussion of Kaplan-Meier re-representation. Next, we discuss the problem of predicting individual Kaplan-Meier curves and formalize the notion of making such predictions using neural networks. The section concludes with a discussion on the different geographical representations explored in this work.

\subsection{Preliminaries}

Let $\{(\bx^{(i)}, e^{(i)}, t^{(i)})\}_{i=1}^{n}$ be a dataset of $n$ instances, where feature vector $\bx^{(i)} \in \mathbb{R}^{m}$, event label $e^{(i)} \in \{0,1\}$, and time of event occurrence $t^{(i)} \in \{0,1,\dots,T\}$. Here, $t^{(i)}$ represents a discrete time at which the event of interest $e^{(i)}$ has occurred (i.e.,~$e^{(i)}=1$) or the last discrete time instance $i$ has been observed and the event has not occurred (i.e.,~$e^{(i)}=0$). In this latter case ($e^{(i)}=0$), when $t^{(i)} = T$ we know the event never occurs to the instance during the study period (spanning $T$ discrete time periods). If, however, $t^{(i)} < T$ then we only know that the instance did not experience the event up to $t^{(i)}$, but don't know what happened during the $T-t^{(i)}$ remaining time. Data having the event-time representation just described are referred to as \textit{censored data}, or more specifically \textit{right-censored} data. An instance $i$ is considered censored when $e^{(i)}=0$ and $t^{(i)} < T$. Censored data, and how we handle them, are elaborated on in a subsequent subsection.

More concretely, $t \in \{1,\dots,T\}$ might represent (as in our experiments) six-month patient follow-up periods, with $t=0$ being the entrance of patients to the study. Entrance to the study, in this case, occurs when a patient is diagnosed with colorectal cancer. For a particular patient $i$, $e^{(i)}=1$ if $i$ dies from colorectal cancer, and $t^{(i)}$ indicates the time of this occurrence. On the other hand, an individual may move across the country, pass away from a non-colorectal cancer related complication or, for whatever reason, lose contact prior to the end of the study period. In such cases (i.e.,~$t^{(i)} < T$), and when patients are not known to have died from their disease, $e^{(i)}=0$.

Each component of instance vector $\bx^{(i)}$ represents the measurement (quantification) of a particular feature. Certain groups of these components will be referenced directly later on in this work and we therefore define notation to reference these particular groups of feature values. Let $\bz$ contain the set of index values that index the geographical features that compose $\bx^{(i)}$ and let $\ba$ denote the full set of index values (i.e.,~ $\ba = \{1,\dots,m\}$). These index sets will be used to reference specific components of $\bx^{(i)}$; i.e.,~$\bx^{(i)}_{\bz}$ is the subvector of instance $i$ containing geographical feature values, and $\bx^{(i)}_{\ba \setminus \bz}$ contains non-geographical feature values.

\begin{table}[h]
\centering
\begin{tabular}{ll}
\toprule
\textbf{Notation} & \textbf{Description} \\ \midrule
$\bx^{(i)} \in \mathbb{R}^m$ & Feature vector of instance $i$. \\
$e^{(i)} \in \{0,1\}$ & Event label of instance $i$. \\
$t^{(i)} \in \{1,\dots,T\}$ & Discrete time of $e^{(i)}$. \\
$\by^{(i)} \in [0,1]^{T}$ & Outcome vector of instance $i$. \\ 
$\hat{\by}^{(i)} \in [0,1]^{T}$ & Predicted outcome vector of instance $i$.\\\midrule
$\bz$ & Set of geographical feature index values.\\
$\ba$ & Set of all feature index values.\\
$\mathcal{M}$ & A map.\\
$\Gamma(\cdot)$ & Function that determines discrete\\
& geographic entity membership.\\\midrule
$P(\cdot)$ & Calculation of a probability.\\
$\mathtt{g}:\mathbb{R}^m \rightarrow [0,1]^{T}$ &  Neural network.\\
$\mathcal{L}(\cdot)$ & An arbitrary loss function.\\
$\mathtt{Smooth}$ & Output smoothing function.\\\midrule
$\pmb{\mathbb{Z}}$ & Adjacency matrix constructed from $\mathcal{M}$. \\
$\mathtt{Common}$& Function that determines whether two geographic entities in\\
&$\mathcal{M}$ are adjacent.\\
$\pmb{Q}_{spec}$& Top $k$ eigenvectors from $\pmb{Q}$, selected based on largest\\
& eigenvalues in $\pmb{\lambda}$.\\
$\mathbf{q}_{label}$& The result of applying kMeans clustering to $\pmb{Q}_{spec}$.\\
$\mathtt{Enrich}$& Function that assigns values in $\pmb{Q}_{spec}$ to an instance.\\\bottomrule
\end{tabular}
\caption{Notation used throughout this work.\label{tab:notation}}
\end{table}

The notation related in this and future sections is related, for convenience, by Table \ref{tab:notation}.

\subsection{Kaplan-Meier Re-representation}

With our preliminary notation defined, we return to elaborating on the censored nature of the data. As mentioned, each instance $i$ has a corresponding event label $e^{(i)}$ and time of event occurrence $t^{(i)}$. We wish, however, to transform this tuple-like representation to one that is in the form of a \textit{Kaplan-Meier survival curve} (KMSC) \cite{kaplan1958nonparametric}. Simply put, a KMSC associates each temporal unit -- in this case the values $1,\dots,T$ -- with a probability of event $e^{(i)}$ not occurring up to that particular time for instance $i$.

Practically speaking, this re-representation will take the form of a vector $\by^{(i)} \in [0,1]^{T}$, where the indices $\tildt \in \{1,\dots,T\}$ denote the temporal units and the entries $\by^{(i)}_{\tildt}$ the corresponding probabilities.

We adopt the re-representation procedure outlined in Chi et al.~\cite{chi2007application} to create $\by^{(i)}$, which can be expressed as
\begin{align}
    \label{eq:rerep}
    y^{(i)}_{\tildt} = \left\{
    \begin{array}{ll}
    1 & \text{if }\tildt < t^{(i)}\\
    0 &\text{if }\tildt \geq t^{(i)} \text{ \& } e^{(i)} = 1\\
    1 - P(e_{\tildt}^{(i)}=1 | e_{\tilde{t}-1}^{(i)}=0) & \text{if }\tilde{t} \geq t^{(i)} \text{ \& } e^{(i)} = 0
    \end{array}
    \right.
\end{align}
where $P(e_{\tildt}^{(i)}=1 | e_{\tildt-1}^{(i)}=0)$ denotes the conditional probability of event $e$ occurring at $\tildt$ given that $e$ has not occurred at $\tildt - 1$. Therefore, for patients whose outcomes are known, $\by^{(i)}$ contains values of 0 and 1 only, whereas a censored patient's vector becomes an estimation of survival probability at the indexical location $\tilde{t} = t^{(i)}$.

\subsection{Predicting Individual KMSC}

Ultimately, the goal of this paper is to learn a hypothesis $\mathtt{g}^* \in \mathcal{G}$, belonging to some [currently] arbitrarily defined hypothesis class $\mathcal{G}$, that most accurately predicts patient-specific KMSCs. Formally, this problem can be written as
\begin{align}
    \mathtt{g}^*=\argmin\limits_{\mathtt{g} \in \mathcal{G}} \left\{ \mathcal{L}\left(\by^{(i)},\mathtt{g}(\bx^{(i)})\right): i=1,\dots,n \right\}
\end{align}
where $\mathcal{L}(\cdot)$ denotes an arbitrary loss function that measures the disparity between the predicted $\by^{(i)}$ (in the future denoted $\hat{\by}^{(i)}$) and the known $\by^{(i)}$. 

In this work, we define our hypothesis class $\mathcal{G}$ to be both shallow and deep neural networks, the specific architecture of which is elaborated on further in this section, with parameterization discussed in the experiments section. We characterize shallow architectures as having one hidden layer and deep architectures as having more than one hidden layer.

\subsubsection{Output Smoothing}

Neural networks are constructed in layer-wise fashion, with each layer consisting of nodes. The inputs are viewed as the first layer, followed by any number of hidden layers. The last of these hidden layers is connected to the output layer. The nature of the output layer is unique to the problem of predicting KMSCs. First, the output nodes are \textit{ordered}. That is, we have a predicted probability for each of the $\tilde{t} = 1,\dots,T$, where $node_{\tilde{t}}^{out}$ is \textit{ordered} before $node_{\tilde{t}+1}^{out}$ because $\tilde{t}$ temporally comes before $\tilde{t}+1$. More importantly, however, the output elicited from these nodes should strictly decrease in temporal order. In other words, we expect $output_{\tilde{t}}^{(i)} \geq output_{\tilde{t}+1}^{(i)}$. Intuitively, even though a patient may have recovered from their disease, one would never expect the probability of survival to go up. However, because the loss function $\mathcal{L}(\cdot)$ typically produces a single value representing the loss across all nodes, the desired strictly decreasing output among temporally ordered output nodes cannot be guaranteed. Therefore, we define a smoothing procedure $\mathtt{Smooth}(\mathbf{output}^{(i)})$, given by
\begin{align}
\label{eq:smooth}
\hat{y}^{(i)}_{\tilde{t}+1} = \min\{output_{\tilde{t}}^{(i)},output_{\tilde{t}+1}^{(i)}\} \text{ for } \tilde{t}=1,\dots,T
\end{align}
which guarantees that the output elicited from the use of a trained model produces strictly decreasing outputs.

\subsection{Geographic Feature Representation}

While we are ultimately concerned with producing a $\mathtt{g}$ that elicits the most accurate predictions, the niche of this work is to:
\begin{enumerate}
\item Show whether geographic-based features improve the quality of predictions.
\item Determine whether a simple binary representation or a richer representation (defined shortly) leads to better predictions.
\item Experimentally quantify the extent of such improvements.
\end{enumerate}
We outline the details of our experiments and data in the next section, where two geographic representations will be explored: a simple binary representation (SBR) and a richer representation produced via spectral analysis (RR-SA).

\subsubsection{Simple Binary Representation}

The simple binary representation (SBR) is a minimalist representation, involving only (a) determination of instance $i$'s discrete geographic entity membership and (b) a binary re-representation of such membership (otherwise referred to as \textit{one hot encoding}), producing a sparse vector with a $1$ in the indexical location corresponding to the geographic entity of which $i$ is a member, and $0$s in all other locations. 

To be as general as possible we assume
that the current geographic features for each instance $i$, denoted $\bx^{(i)}_{\bz}$, can be used to obtain the single discrete geographic unit of which $i$ is a member. As an example, in our experiments, we use ZCTA (zipcode tabulation area) as our discrete geographic unit.

To formalize the notion of eliciting discrete geographic unit membership, let 
\begin{align}
\label{eq:geomem}
    x^{(i)}_{b} = \Gamma(\bx^{(i)}_{\bz},\mathcal{M})
\end{align}
where $\Gamma(\cdot)$ is a function that transforms the geographic feature values of instance $i$ to an ID value, denoted $x^{(i)}_{b}$, representing the single geography entity in a map $\mathcal{M}$ (defined shortly) that $i$ is a member of. Depending upon the geographic information encapsulated by $\bx^{(i)}_{\bz}$, the function $\Gamma(\cdot)$ and map $\mathcal{M}$ may take on different forms. 

In this work our geographic features are (lat,lon) coordinate pairs. Therefore, we provide a specific definition (Definition \ref{def:map}) outlining the map $\mathcal{M}$ that makes use of (lat,lon)-specified geography.

\begin{definition}
\label{def:map}
Define $\mathcal{M}$ to be a \textbf{map}, given by
\begin{align}
    \mathcal{M} = \left \{(key_l,value_l)\right\}_{l=1}^{p}
\end{align}
where $key_l$ is the unique postal code of geographic unit $l$ and $value_l$ is an ordered set of (lat,lon) coordinate pairs denoting the bounding geographic region of $l$.

Map $\mathcal{M}$ is a continuous geographic region, characterized by
\begin{align}
    \left\{ \forall \mathtt{l} \exists \mathtt{l}^{\prime}: value_{\mathtt{l}}^{q} = value_{\mathtt{l}^{\prime}}^{j} \text{ for } \mathtt{l},\mathtt{l}^{\prime} \in \{1,\dots,p\} \text{ \& } \mathtt{l} \neq \mathtt{l}^{\prime}
    \right\}
\end{align}
where $value_{\mathtt{l}}^{q} = value_{\mathtt{l}^{\prime}}^{j} \triangleq (lat_{\mathtt{l}}^q = lat_{\mathtt{l}^{\prime}}^j) \cap (lon_{\mathtt{l}}^q = lon_{\mathtt{l}^{\prime}}^j$).
\end{definition}


Given our definition of $\mathcal{M}, \Gamma(\cdot)$ is a function that determines whether a point, given by $\bx^{(i)}_{\bz}$, is on the interior of each ZCTA in $\mathcal{M}$. When the ZCTA having $\bx^{(i)}_{\bz}$ in the interior is found, $x^{(i)}_{b}$ is set equal to the ZCTA's identifier. A binarization procedure (one hot encoding), denoted $\mathtt{Bin}$, is applied to  $x^{(i)}_{b}$, thus producing a sparse vector representation.

\begin{figure}[h]
    \centering
    \includegraphics[scale=.60]{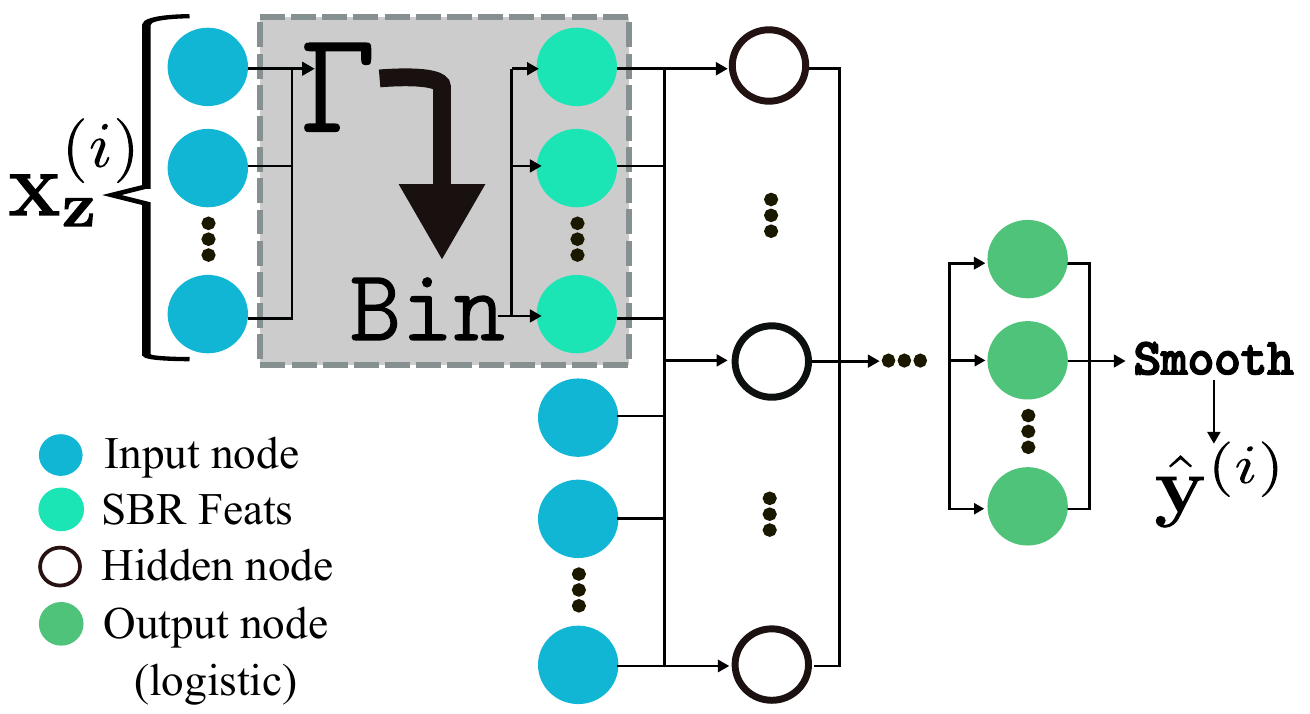}
    \caption{SBR neural network architecture.}
    \label{fig:sbr-arch}
\end{figure}

%
%
%

Figure \ref{fig:sbr-arch} illustrates the network architecture using the SBR methodology.

While we expect the addition of SBR features to elicit a hypothesis having some predictive performance improvement over a hypothesis employing only non-geographic features, richer representations that better capture the continuous nature of the defined  geographic region hold greater promise.

\subsubsection{Learning a Rich Geographic Representation}
To obtain richer geographic features, we adopt a spectral clustering-based approach (spectral analysis) to geographical feature re-representation. At a high level, this method first computes the geographic adjacency of the discrete entities that comprise $\mathcal{M}$, thus producing an adjacency matrix. Spectral analysis is then performed on this matrix. Spectral analysis involves first solving for the eigenvalues and eigenvectors of the adjacency matrix. Second, the top (i.e.,~largest) $k$ eigenvalues are used to select the top $k$ corresponding eigenvectors, forming a $p \times k$ matrix. The $p$ rows correspond to the $p$ geographic entities (one row corresponds to one of the $p$ geographic entities). The $k$ values associated with each entity are then used as predictive input features.  

To express this procedure more formally, let $\pmb{\mathbb{Z}} = \mathtt{Adj}\left(\mathcal{M}\right)$ denote the adjacency (i.e.,~affinity, similarity) matrix, where the $l,v$-th entry corresponds to the geographic adjacency relationship between the $l$-th and $v$-th discrete geographic entities, which is given by
\begin{align}
    \label{eq:adjmatform}
    [\pmb{\mathbb{Z}}]_{l,v} = \left\{
    \begin{array}{ll}
    1 & \text{if }\ \mathtt{Common}(values_l,values_v) = True \\
    & \text{\& } l \neq v\\
    0 & \text{otherwise}
    \end{array}
    \right.
\end{align}
where the function $\mathtt{Common}(\cdot)$ evaluates whether $values_l$ and $values_v$ share a common element. In the context of the $\mathcal{M}$ described by Definition \ref{def:map}, $\mathtt{Common}(\cdot)$ determines whether or not $values_l$ and $values_v$ have at least one coordinate pair in common.

Spectral clustering is performed by doing $\mathbf{q}_{label}= kMeans \left(\pmb{Q}_{spec}\right)$, where $kMeans(\cdot)$ assigns one of $k$ cluster labels to each of the $p$ column elements using the $k$-means clustering algorithm, and where
\begin{align}
\label{eq:qspec}
\pmb{Q}_{spec} = \mathtt{Top}_k\left(\pmb{Q},\pmb{\lambda}\right).
\end{align}
The function $\mathtt{Top}_k(\cdot)$ finds the largest values in $\pmb{\lambda}$, selects the corresponding columns in $\pmb{Q}$, and forms the $\pmb{Q}_{spec} \in \mathbb{R}^{k\times p}$ submatrix. The matrix $\pmb{Q}$, composed of eigenvectors, and vector $\pmb{\lambda}$, composed of eigenvalues, are obtained by solving the system of equations given by 
\begin{align}
\label{eq:sceiq}
\pmb{\mathbb{Z}}\pmb{Q} = \pmb{\lambda} \pmb{Q} .
\end{align}
Practically speaking, the column-wise elements of $\pmb{Q}_{spec}$ are used as $k$ geographical features when learning $\mathtt{g}$ -- this is spectral \textit{analysis} -- and the labels $\mathbf{q}_{label}$ are used for visualization purposes (as in our experiments in the next section) -- this is spectral \textit{clustering}. In other words, instead of using a [necessarily] binarized form of the label assignment elicited from $k$-means clustering as features, we use the eigenvectors [on which clustering is performed], which preserves cluster composition. 

To further differentiate spectral clustering from spectral analysis, we detail the spectral clustering procedure in Algorithm \ref{algo:sc}. Omission of the final line, highlighted in red, yields the spectral analysis procedure used to create the rich representation.

\begin{algorithm}
    \caption{Spectral Clustering\label{algo:sc}}
    \begin{algorithmic}[1]
    \STATE Obtain adjacency matrix $\pmb{\mathbb{Z}}$ using \eqref{eq:adjmatform}.
    \STATE Solve \eqref{eq:sceiq} for $\pmb{Q}$ and  $\pmb{\lambda}$.
    \STATE Obtain $\pmb{Q}_{spec}$ as outlined in \eqref{eq:qspec}.
    \STATE \textcolor{red}{Apply kMeans clustering to }$\color{red}\pmb{Q}_{spec}$ \textcolor{red}{to obtain }$\color{red}\mathbf{q}_{label}$\textcolor{red}{.}
    \end{algorithmic}
\end{algorithm}

In other words, spectral analysis is a sub-procedure of spectral clustering, wherein the \textit{clustering} step is omitted.

Finally, when an instance $\bx$ is encountered, a procedure $\mathtt{Enrich}(\bx_{\bz},\mathcal{M},\pmb{Q}_{spec})$ is called that obtains the $k$-valued column of $\pmb{Q}_{spec}$ that corresponds to the particular geographic entity that $\bx$ belongs. $\mathtt{Enrich}$ is outlined in Algorithm \ref{algo:Enrich}.
\begin{algorithm}
	\caption{Enrich Geographic Features $\pmb{\mathtt{Enrich}}$}
	\label{algo:Enrich}
	\begin{algorithmic}[1]
	    \REQUIRE $\bx_{\bz},\mathcal{M},\pmb{Q}_{spec}$
	    \STATE $x_b = \Gamma(\bx_{\bz},\mathcal{M})$ From \eqref{eq:geomem}.
	    \STATE Using $x_b$ find the $l$ such that $x_b = key_l: l \in \{1,\dots,p\}$.
	    \ENSURE Return column vector $[\pmb{Q}_{spec}]_l$
	\end{algorithmic}
\end{algorithm}

\begin{figure}[h]
    \centering
    \includegraphics[scale=.60]{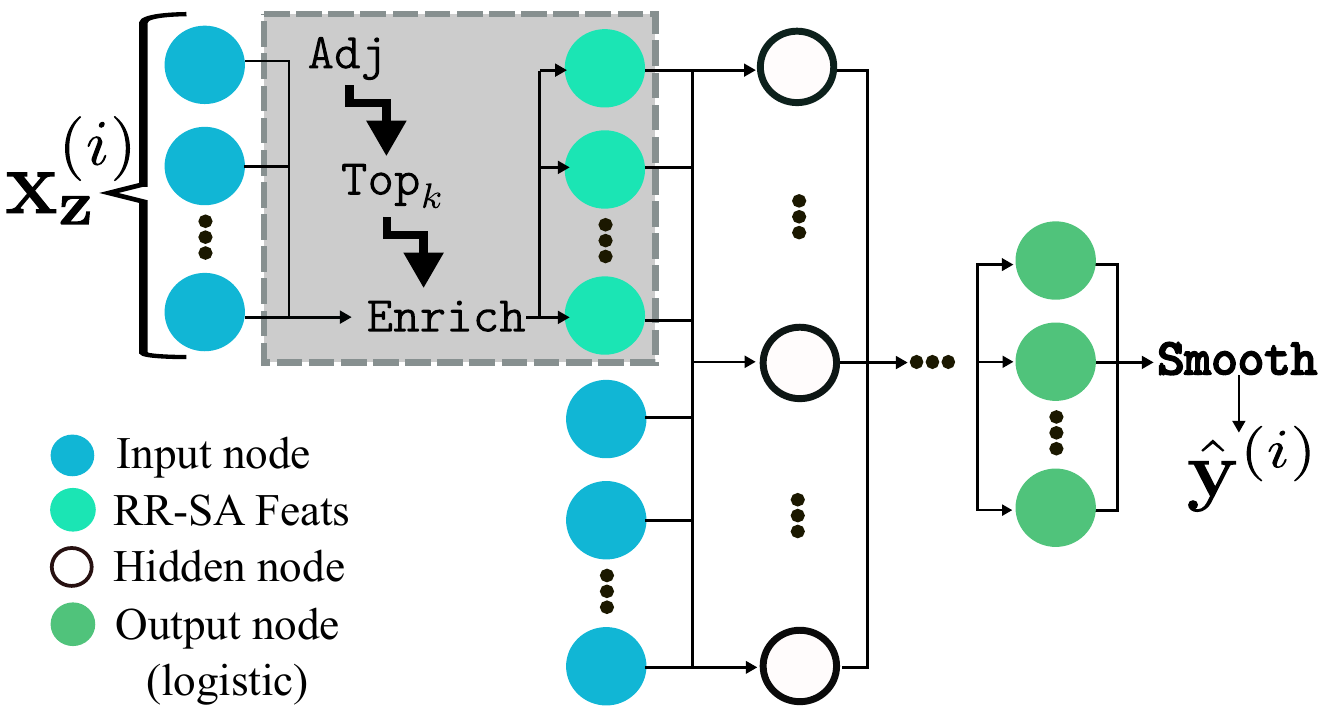}
    \caption{RR-SA neural network architecture.}
    \label{fig:rr-sc-arch}
\end{figure}

The network architecture that encapsulates the spectral analysis process is shown in Figure \ref{fig:rr-sc-arch}\footnote{In our experiments $\mathbf{x}_{\mathbf{z}^(i)}$ are latitude and longitude coordinates.}.

\section{Predicting Colorectal Cancer Survival}

In this section we begin by providing an in-depth description of the data used in our experiments, followed by an outline of the technical details of our experiments. Subsequently, we discuss experiments and results comparing average predicted survival curve against average actual survival curve by model, as well as mean absolute error by model when the smoothing procedure is removed.

\subsection{Colorectal Cancer Survival Data for the State of Iowa}

Our data were provided by the Iowa Cancer Registry (ICR), State Health Registry of Iowa (SHRI), and the Iowa Department of Public Health (IDPH). Each instance represents a patient who has been diagnosed with colorectal cancer and whose residence at the time of diagnosis is in the state of Iowa. The dataset consists of $n=46116$ patients and, initially, $m=71$ features. After removing identifiers and features having a large number of instances with missing values (\% missing $>$ 50\%), we were left with $m=26$ distinct features (including unprocessed geographic coordinates). After binarizing discrete features, $m=386$ (excluding geographic features). When using SBR geographical re-representation, $m=1364$ ($386$ non geographic features and $p=978$ binarized geographic features), and $m=386 + k$ when using the RR-SA geographic representation (where $k$ is parameterized and therefore user-dependent). When the Kaplan-Meier re-representation is applied to the dataset, we obtain $\by^{(i)}$ vectors having $T=53$ elements, where each element represents the patient's current vital status (alive$=1$ or dead$=0$), or a probability of survival when an instance becomes censored, as described by \eqref{eq:rerep}. Each $\tilde{t} \in \{1,\dots,53\}$ represents six months.

\begin{figure}[h]
    \centering
    \includegraphics[scale=0.045]{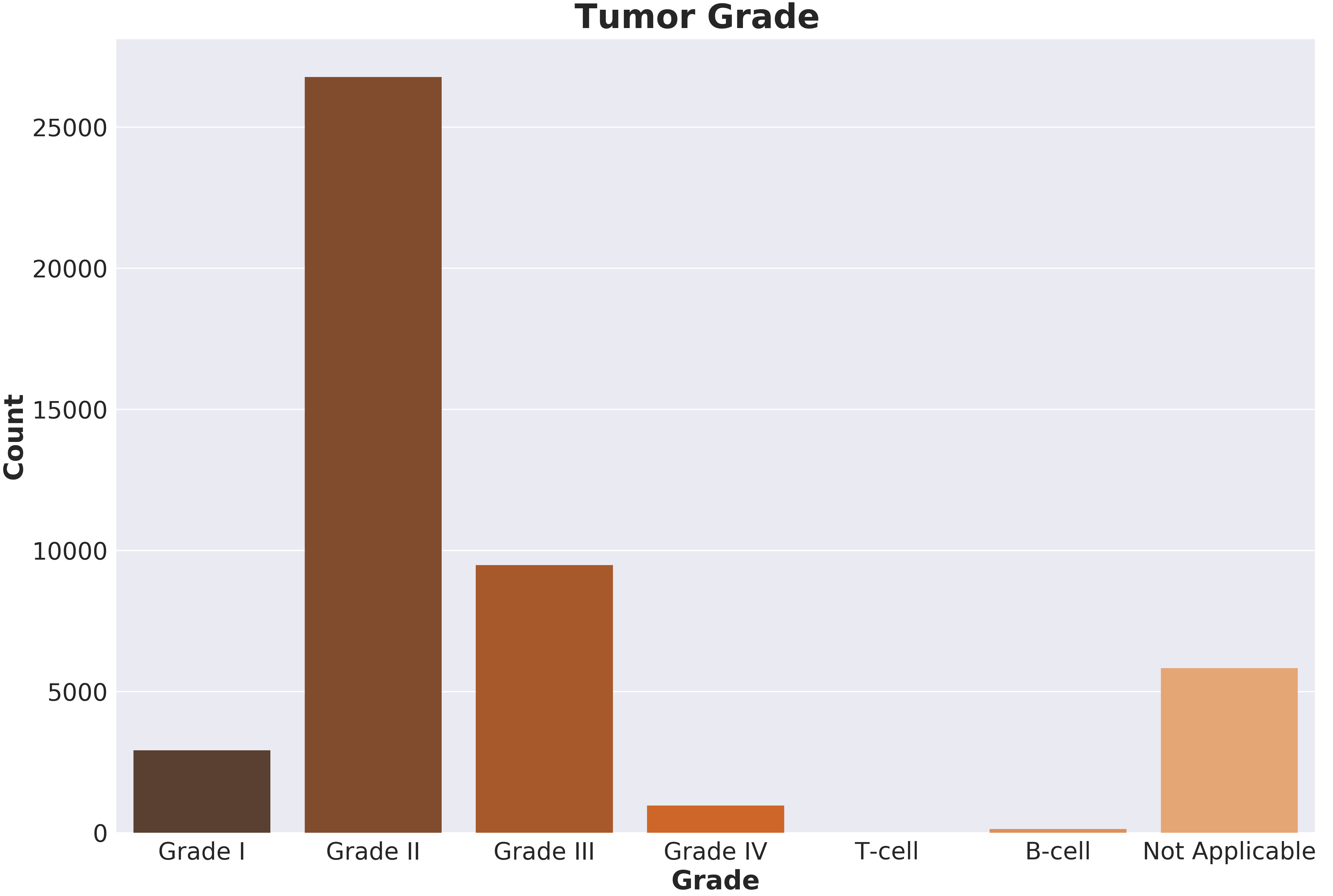}
    \caption{Tumor grade at diagnosis for patients in the state of Iowa: Years 1989 to 2012.\label{fig:grade}}
\end{figure}

\begin{figure}[h]
    \centering
    \includegraphics[scale=0.045]{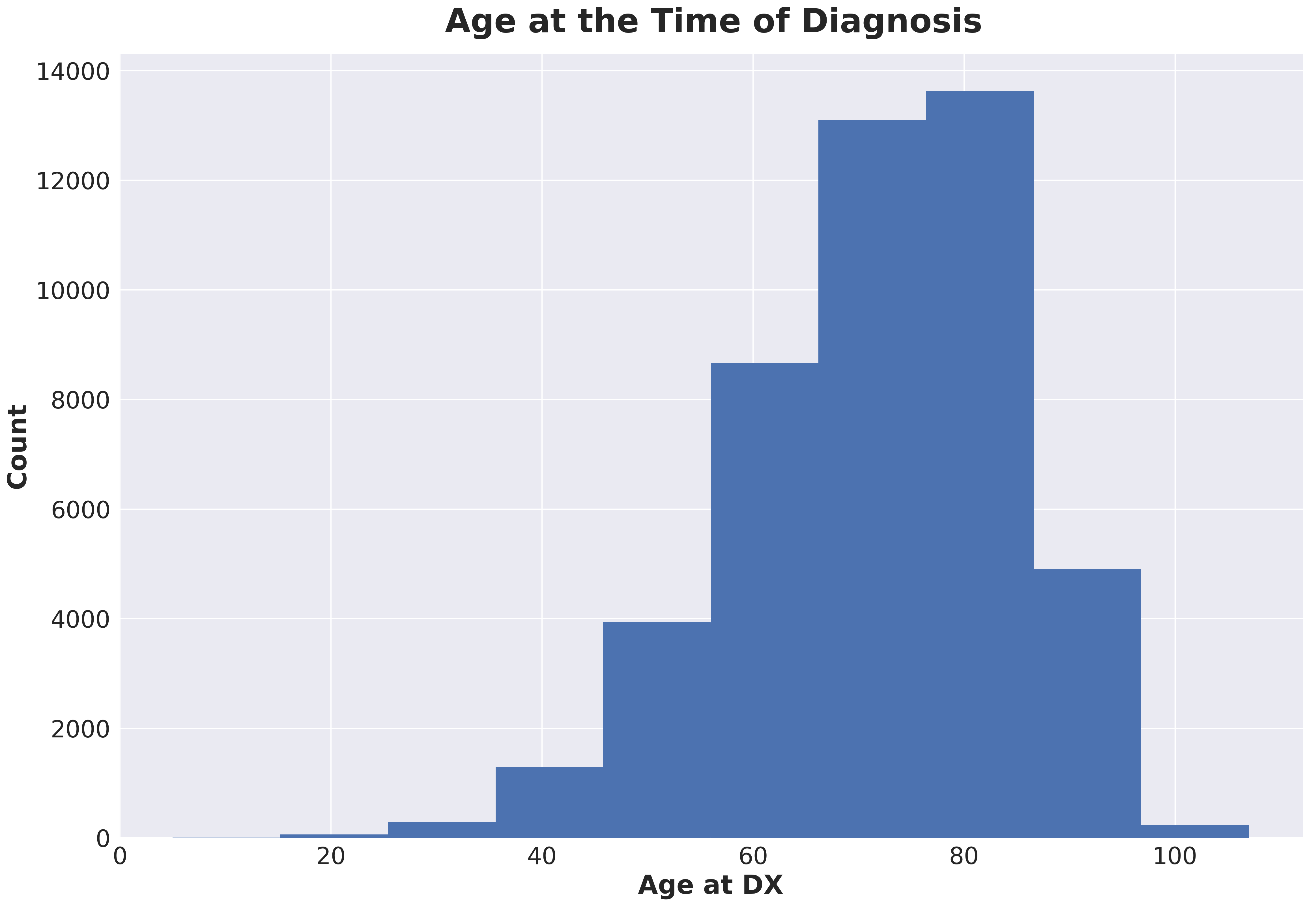}
    \caption{Age of colorectal cancer diagnosis for patients in the state of Iowa: Years 1989 to 2012.\label{fig:age}}
\end{figure}

The $24$ distinct non-geographic features pertain to various patient-specific characteristics, which can be categorized as \textit{disease-based} and \textit{demographic-based}. Disease-based features include tumor grade, tumor histology and tumor marker; we show a histogram of tumor grade in Figure \ref{fig:grade}. Demographic-based features include marital status, race, and age at diagnosis; we show a histogram of age at diagnosis in Figure \ref{fig:age}. These selected features (age and tumor grade) have been shown to be indicative of not receiving timely cancer treatment \cite{ward2013does}, which we believe will help in predicting cancer survival, although analysis of such factors is beyond the scope of this work.

\subsection{Predictive Setting, Pamaterization and Results}

As outlined in the introduction, we wish to address the following:
\begin{enumerate}
    \item On average, can colorectal cancer survival curves be
reasonably predicted for patients in the state of Iowa?
    \item Do geographic features improve the quality of predicted
colorectal cancer survival curves for patients in the state
of Iowa?
    \item Do richer geographical feature representations improve predictive
performance more than simpler representations?
\end{enumerate}

To such an end, we propose to use $10$-fold validation where, for each fold, we find a $\mathtt{g}^*$ for each of the following types of model:
\begin{enumerate}[label=(\roman*)]
\item A model constructed using no geographical features (No Geo).
\item A model constructed using SBR-derived geographical features, as outlined by Figure \ref{fig:sbr-arch} (SBR).
\item Models constructed using RR-SA-derived geographical features, as outlined by Figure \ref{fig:rr-sc-arch}, where the values $k=10,20,30,40$ will be explored (RR-SA).
\end{enumerate}

We then examine two different factors:
\begin{enumerate}[label=(\alph*)]
\item Each model's average survival curve prediction on the test set, taken over the $10$ folds, as compared to the actual average survival curve, taken over all $\by^{(i)}$. We devise a metric we term \textit{area between curves} (ABC) that measures the area-wise disparity between the two curves.
\item Each model's mean absolute error in the absence of the output smoothing procedure (described in Section 2.C.1).
\end{enumerate}

\subsubsection{Model Parameterization}

Our models are constructed using Tensorflow, employing fully connected layers, trained using sigmoidal cross entropy as the loss function $\mathcal{L}(\cdot)$. The logistic activation function is used for all nodes. Each model is trained using a maximum of $2500$ epochs with a $15\%$ batch size. While the connectedness of the architecture, activation function, epochs, and batch size are all tunable parameters, we elect to focus on finding the optimal number of hidden layers and corresponding hidden nodes for each layer. Table \ref{tab:params} shows the average optimal architecture for each of the models, taken over the 10 folds.

\begin{table}[t]
    \centering
    \begin{tabular}{ll}
    \toprule
    Model & Avg Optimal Architecture \\\midrule
    No Geo & 1.5:[83,30] \\
    SBR & 1.9:[260,122] \\
    RR-SA, $k=10$ & 1.5:[82,36] \\
    RR-SA, $k=20$ & 1.5:[102,44] \\
    RR-SA, $k=30$ & 1.6:[87,45] \\
    RR-SA, $k=40$ & 1.5:[80,44] \\\bottomrule
    \end{tabular}
    \caption{Average optimal architecture by model over the 10 folds (e.g., No geo had 1.5 hidden layers, on average, where the first layer had 83 nodes , on average, and the second layer had 30 nodes, on average).\label{tab:params}} 
\end{table}

\begin{figure*}[h]
    \centering
    \begin{subfigure}[]{.32\linewidth}
        \centering
        \includegraphics[scale=.28]{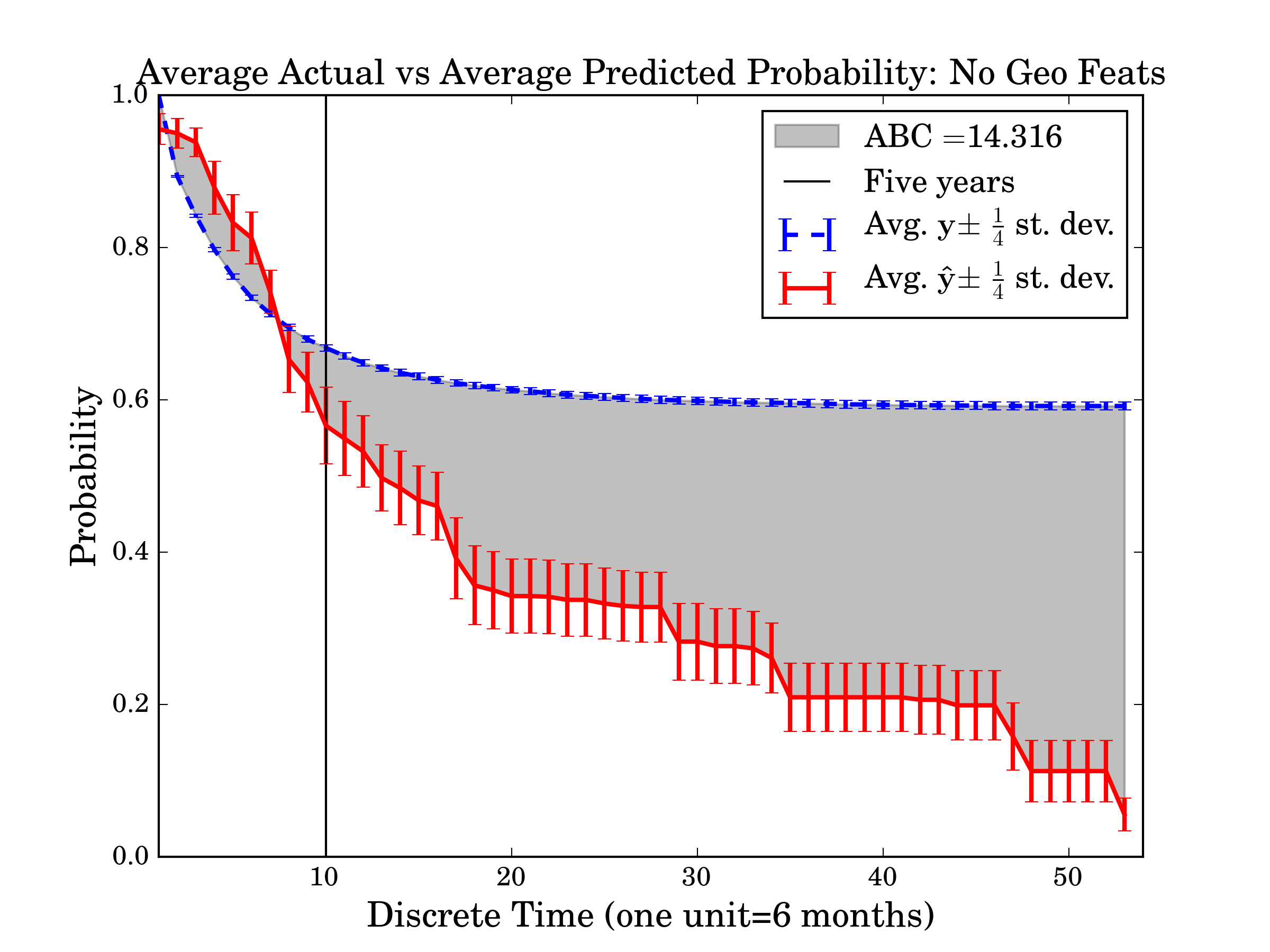}
        \caption{No geo feats (MAE = 0.467).\label{fig:scnogeo}}
    \end{subfigure}
    \begin{subfigure}[]{.32\linewidth}
        \centering
        \includegraphics[scale=.28]{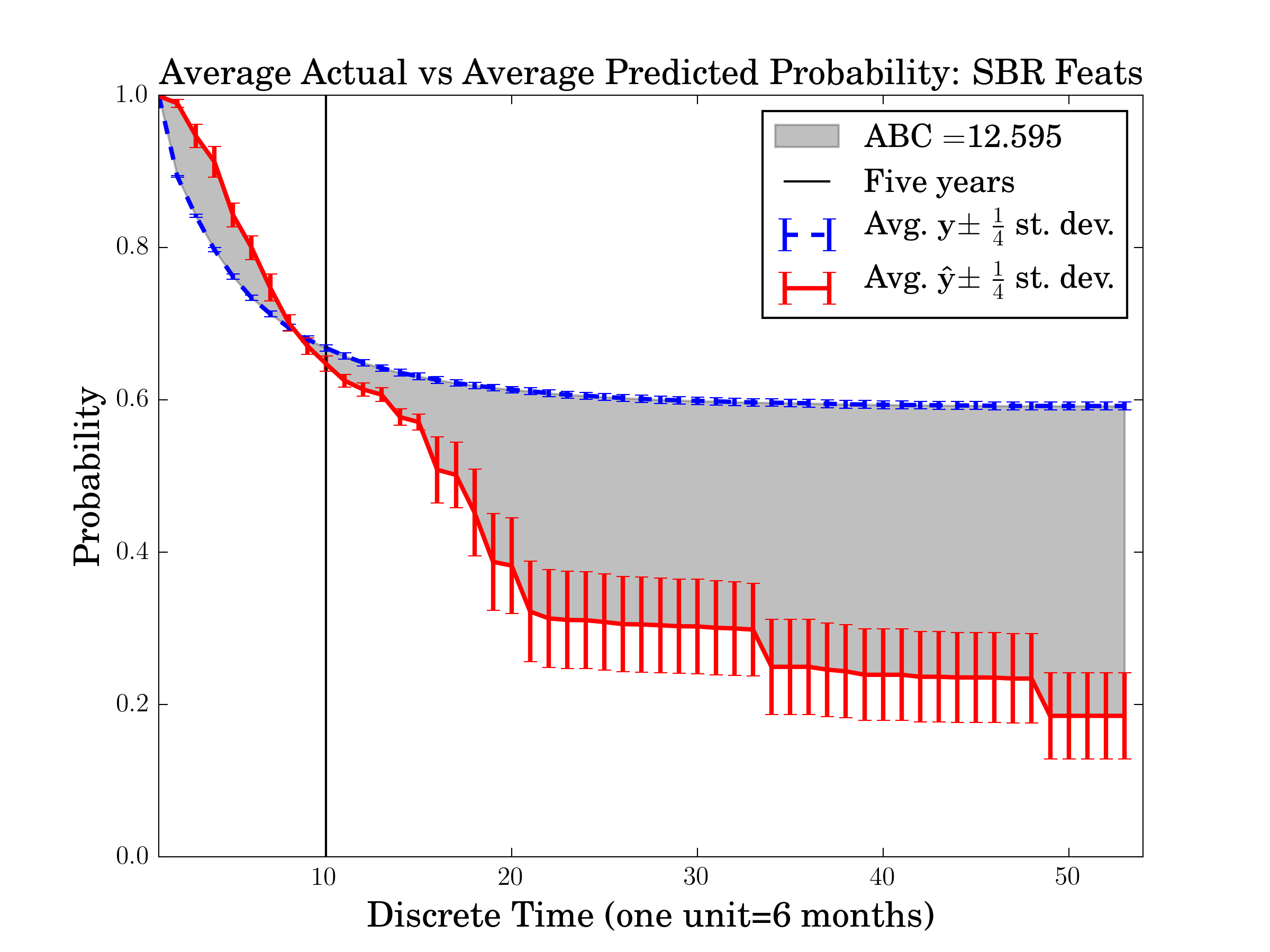}
        \caption{SBR (MAE = 0.4512).\label{fig:scwithgeo}}
    \end{subfigure}
    \begin{subfigure}[]{.32\linewidth}
        \centering
        \includegraphics[scale=.28]{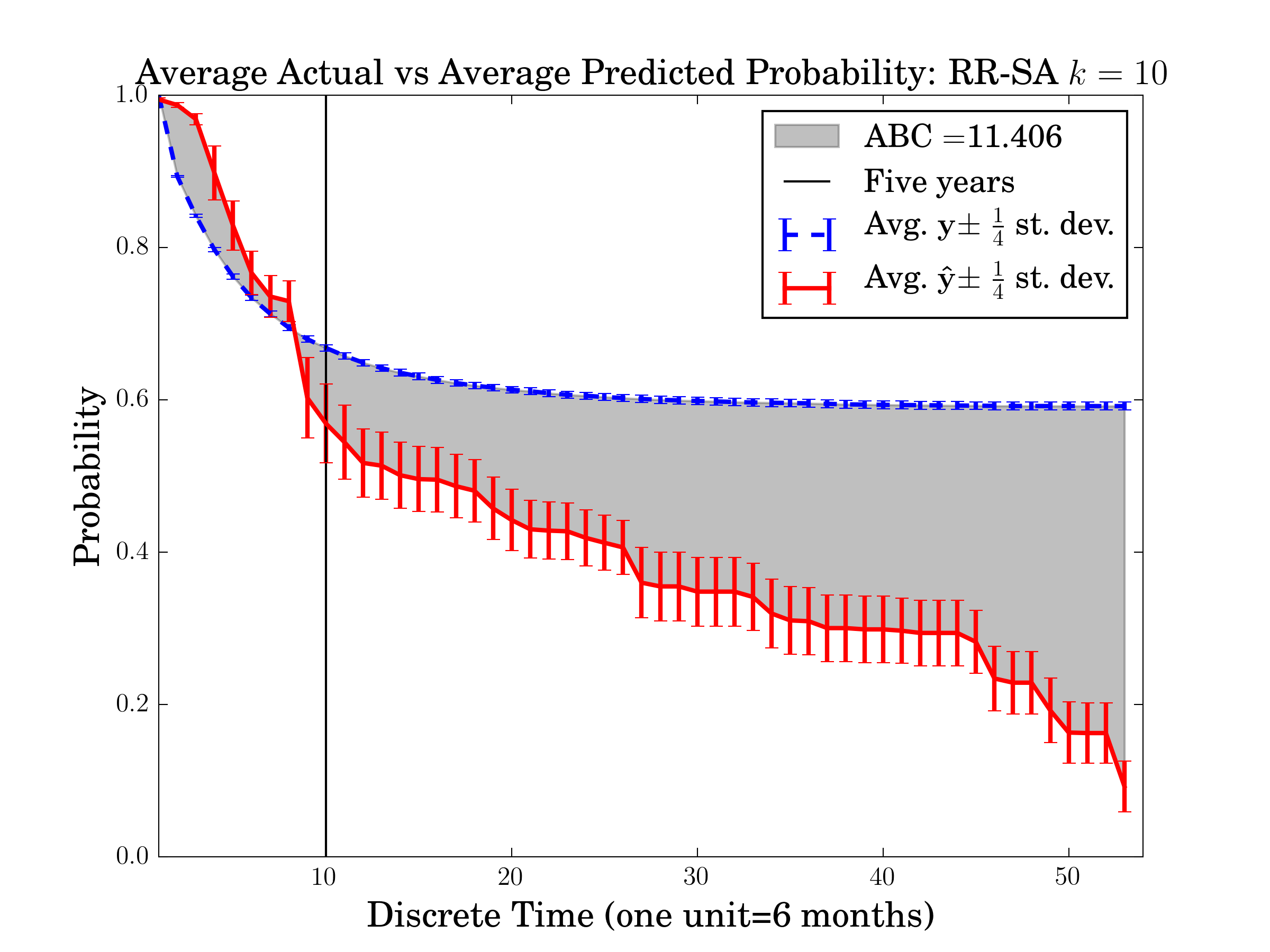}
        \caption{RR-SA, $k=10$ (MAE = 0.446).\label{fig:sc10geo}}
    \end{subfigure}\par 
    \begin{subfigure}[]{.32\linewidth}
        \centering
        \includegraphics[scale=.28]{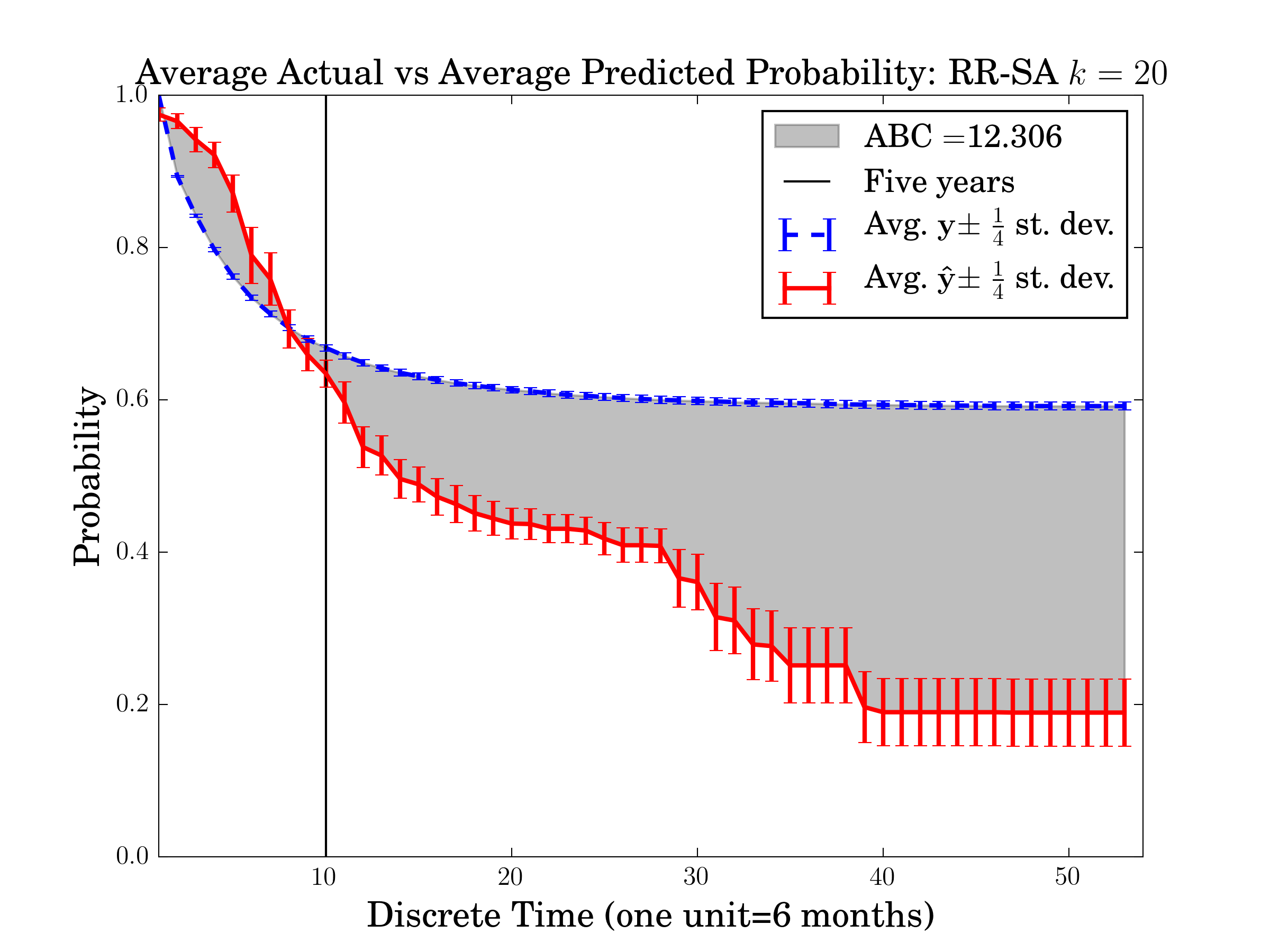}
        \caption{RR-SA, $k=20$ (MAE = 0.453).\label{fig:sc20geo}}
        \label{}
    \end{subfigure}
    \begin{subfigure}[]{.32\linewidth}
        \centering
        \includegraphics[scale=.28]{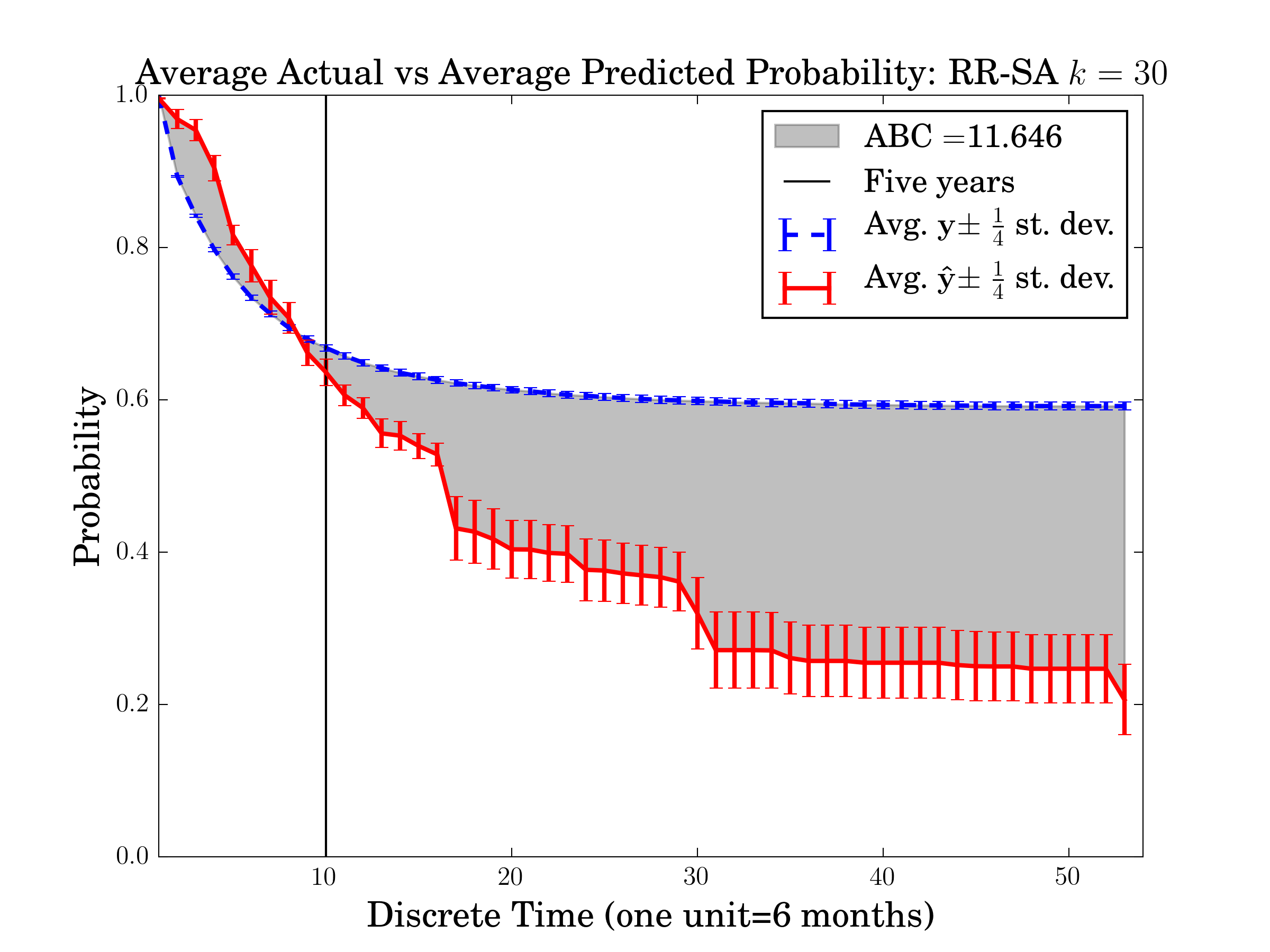}
        \caption{RR-SA, $k=30$ (MAE = 0.445).\label{fig:sc30geo}}
    \end{subfigure}
    \begin{subfigure}[]{.32\linewidth}
        \centering
        \includegraphics[scale=.28]{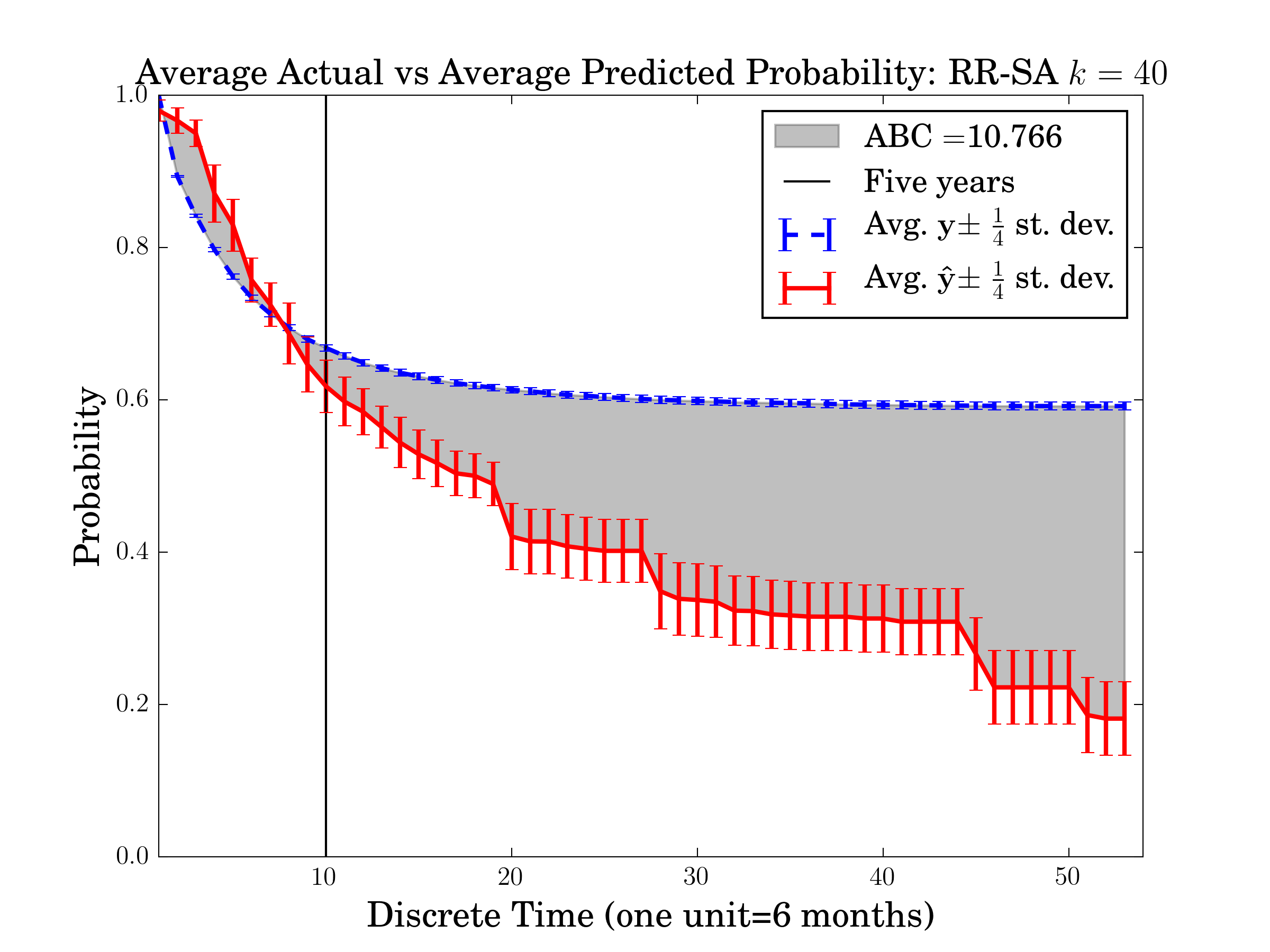}
        \caption{RR-SA, $k=40$ (MAE = 0.442).\label{fig:sc40geo}}
    \end{subfigure}
    \caption{Actual vs. Predicted.\label{fig:avpsc}}
\end{figure*}

In Table \ref{tab:params} we can see that, on average, the optimal architecture is relatively comparable among all models with the exception of SBR (and to a degree RR-SA, $k=20$). First, this suggests that the addition of rich geographic features, as defined in this work (obtained using spectral analysis), do not affect the architectural complexity of the model. However, SBR seems to significantly increase such complexity. This is somewhat expected, as SBR is represented as a large, sparse vector, which can be contrasted with the comparatively small vector of RR-SA.

\subsubsection{Average Actual vs Average Predicted Survival}

The results comparing the average actual survival curve against the average predicted survival curve, by model, are presented in Figure \ref{fig:avpsc}. Henceforth, these curves will simply be referred to as \textit{actual} and \textit{predicted}. In these figures we also shade the region between the actual and predicted curves and provide a value representing the total area covered by this region. We will use this value, henceforth referred to as \textit{area between the curves} (ABC for short), as a means of comparing the predictive quality of the six different models (where lower ABC is better). We also include the mean absolute error for each model, reported as an average over the 53 outputs.

Comparing Figure \ref{fig:scnogeo} with Figures \ref{fig:scwithgeo} through \ref{fig:sc40geo} we first see that the addition of geographical features has uniformly improved the quality of the predictions, on average, as can be observed visually and by comparing ABC values. The MAE values in parenthesis support this conclusion.

Secondly, comparing Figure \ref{fig:scwithgeo} with Figures \ref{fig:sc10geo} through \ref{fig:sc40geo}, we observe that models using richer geographical representations (RR-SA) perform better (\ref{fig:sc10geo} - \ref{fig:sc40geo}) than a model trained using a simple representation (\ref{fig:scwithgeo}), again in terms of both ABC and MAE.

\begin{figure*}[t]
    \centering
    \begin{subfigure}[]{.24\linewidth}
        \includegraphics[scale=.035]{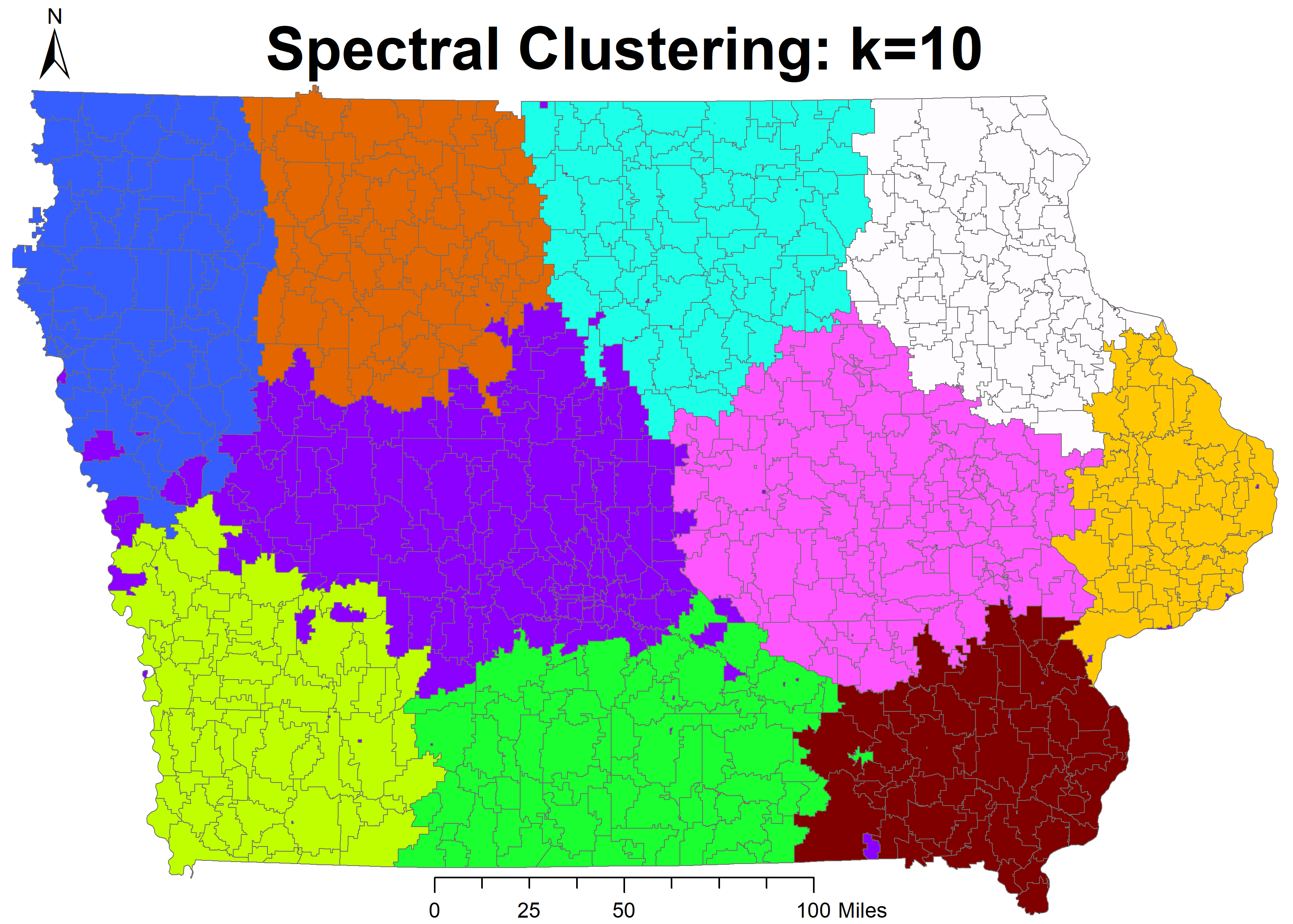}
    \end{subfigure}
    \begin{subfigure}[]{.24\linewidth}
        \includegraphics[scale=.035]{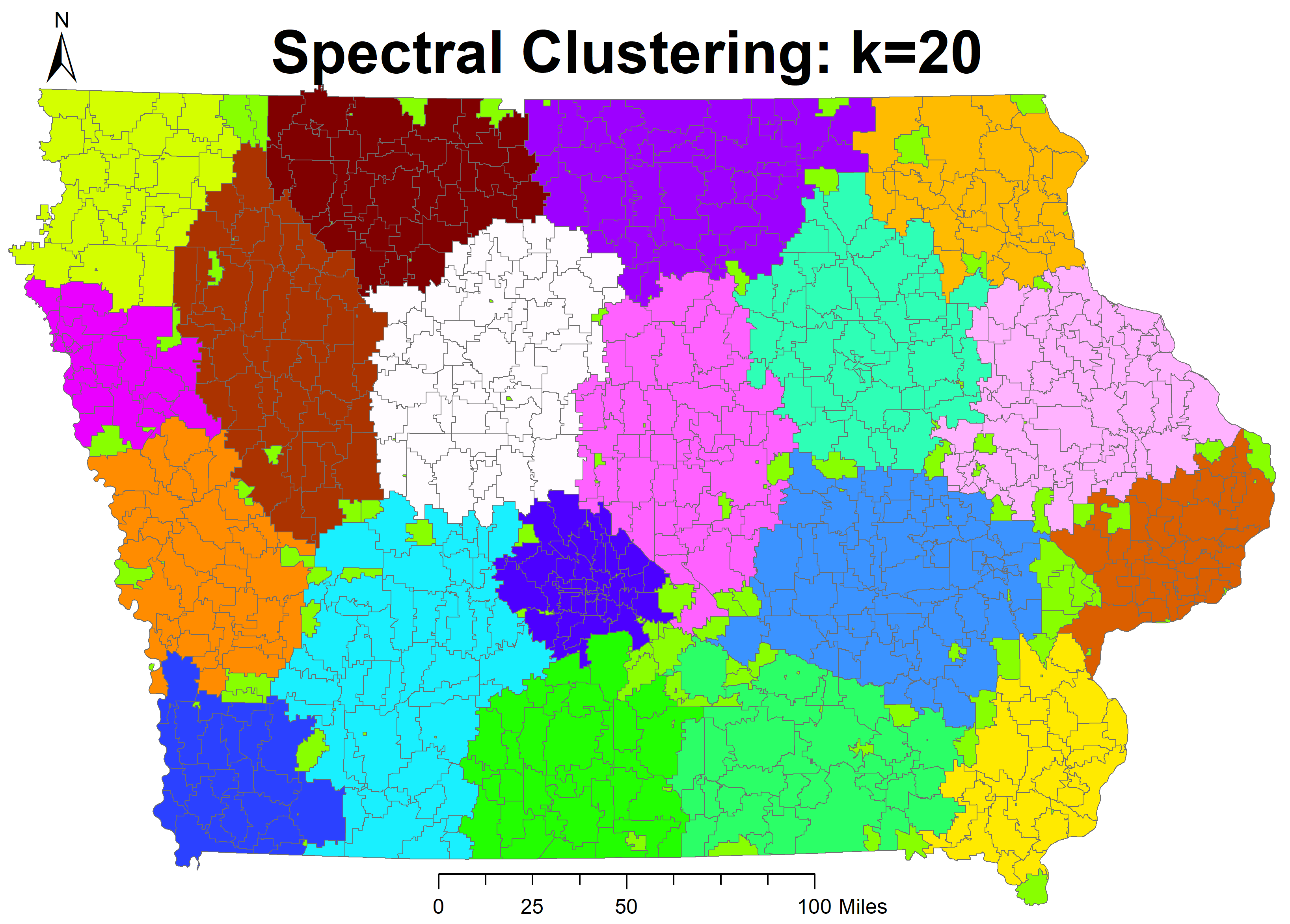}
    \end{subfigure}
    \begin{subfigure}[]{.24\linewidth}
        \includegraphics[scale=.035]{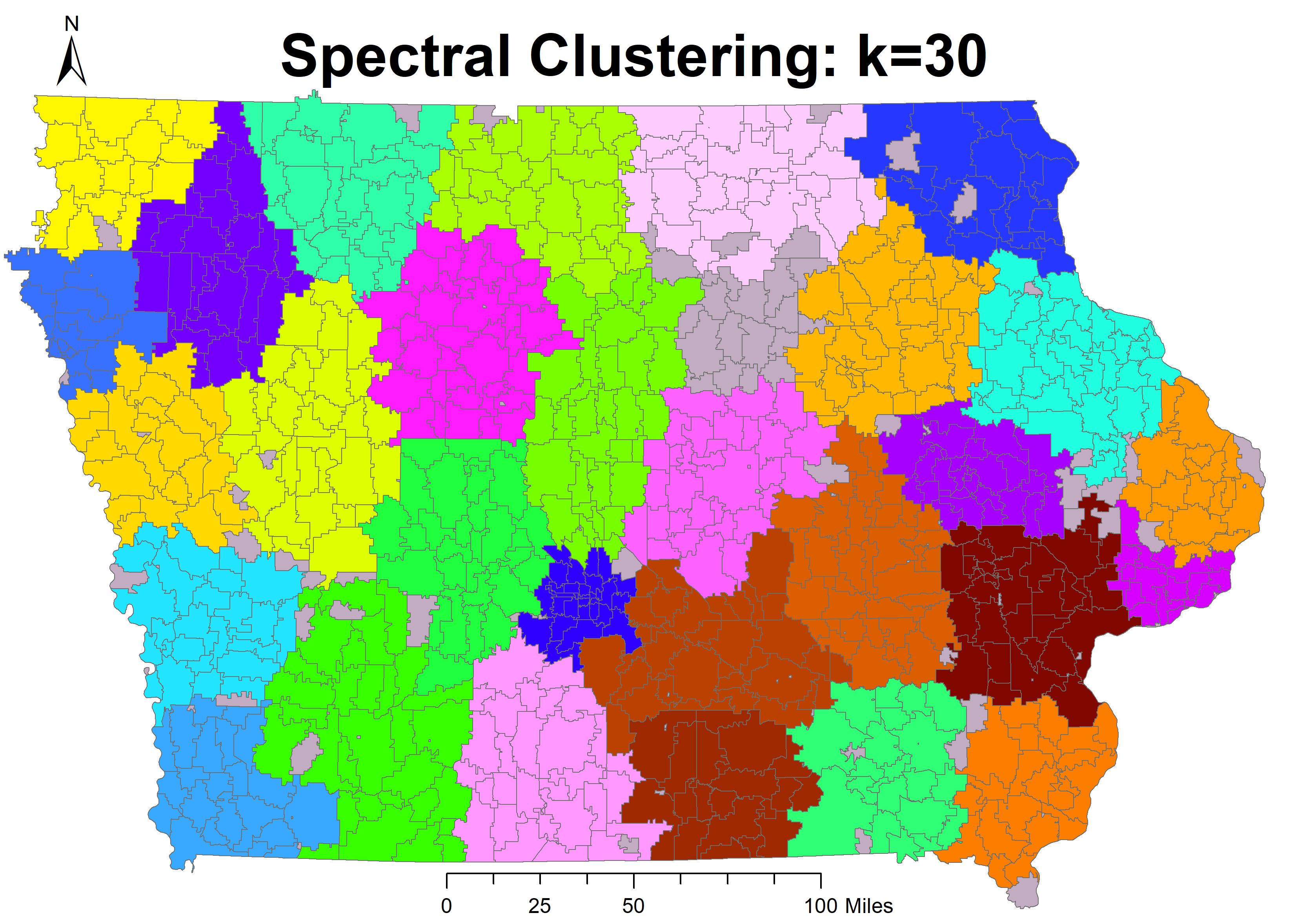}
    \end{subfigure}
    \begin{subfigure}[]{.24\linewidth}
        \includegraphics[scale=.035]{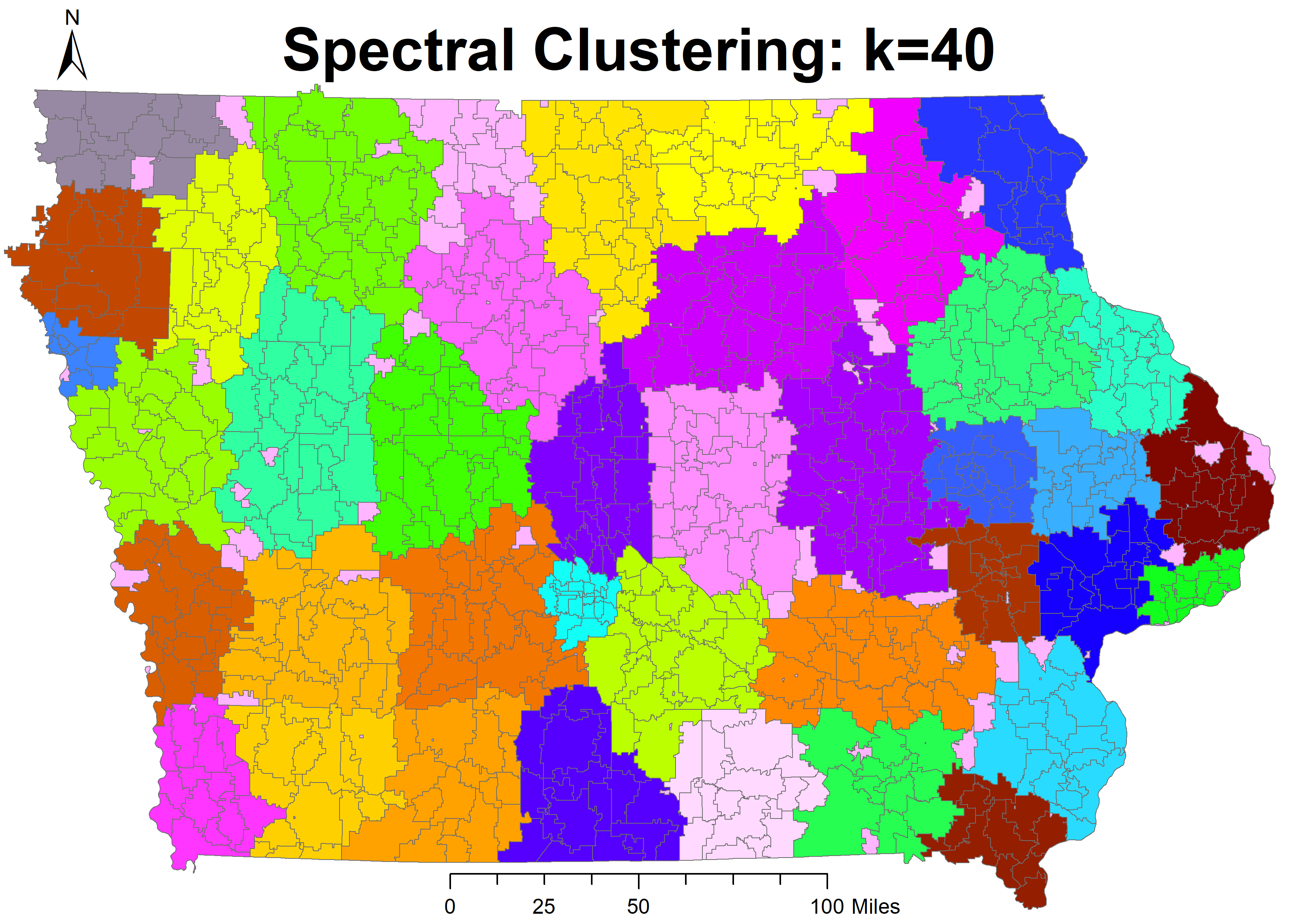}
    \end{subfigure}
    \caption{Spectral clustering results for $k=10,20,30,40$, where color denotes cluster membership.}
    \label{fig:sc_map}
\end{figure*}

However, there are also RR-SA model performance differences depending on the parameterized $k$ value. Interestingly, there seems to exist a non-linear relationship between $k$ and performance, with $k=10$ outperforming $k=20$, and $k=30$ outperforming $k=10$; $k=40$ performs the best out of all models. We believe this nonlinear relationship may be accounted for by the fact that higher values of $k$ lead to more localized models, yet can also produce sparse, disjointed clusters. This point is supported by our clustering visualizations reported in Figure \ref{fig:sc_map} and discussed in Section III.B.4.

In examining the different predicted survival curves we have a few observations, summarized as follows. First, we observe that predictive performance increases are mostly realized after the five-year mark. This is, on one hand, intuitive because predicting survival at times closer to the diagnosis is easier than predicting survival at later times. On the other hand, noticeable deviation of the predicted curves uniformly occurs across all models at or around this five-year mark. Therefore, model improvement wrought by using richer geographical representations is realized, by-in-large, at times beyond the five-year mark. Explanation as to \textit{why} such a deviation is present in all models requires further investigation beyond the scope of this work. 

In summary, we find that
\begin{enumerate}
    \item On average, colorectal cancer survival curves can be
reasonably predicted for patients in the state of Iowa.
    \item Geographic features do improve the quality of predicted
colorectal cancer survival curves for patients in the state
of Iowa by 25\% (on average).
    \item On average, richer geographical feature representations improve predictive performance by 15\% over simpler representations.
\end{enumerate}

\begin{figure}[h]
    \centering
    \includegraphics[scale = 0.05]{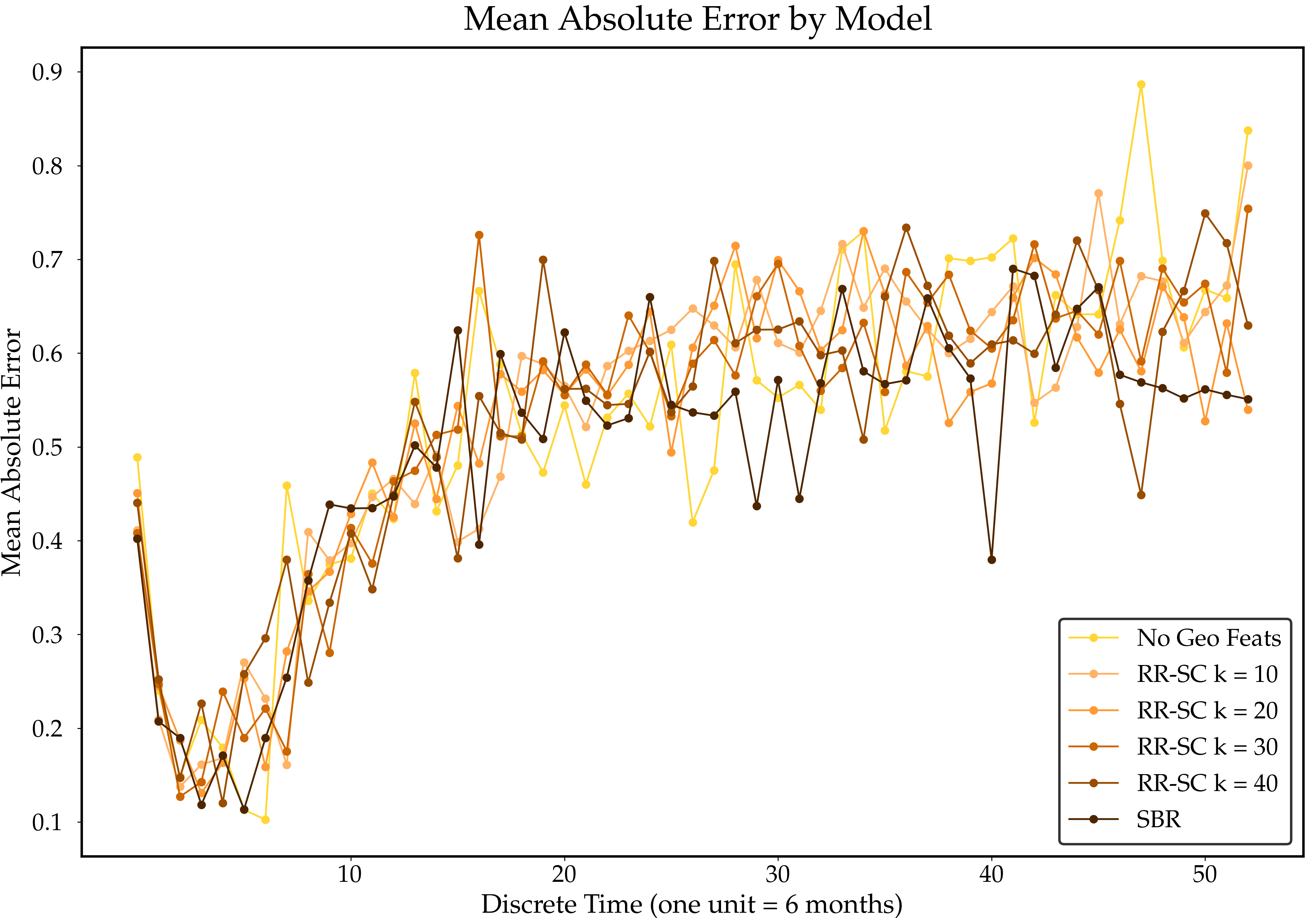}
    \caption{MAE by model.\label{fig:moderr}}
\end{figure}

\subsubsection{Model Errors}

To further examine model performance we compare the mean absolute error of each of the models, measured at each time unit. The results comparing average error by model type are presented in Figure \ref{fig:moderr}. Note that we report these error results without using the post-processing technique described in Section II.C.1 (output smoothing). We do this to provide a slightly different look at model performance over the result presented in Figure \ref{fig:avpsc}.

First, it is clear that all models seem to follow a similar pattern in terms of the observed error by output node, which is also found when comparing average model output in Figure \ref{fig:avpsc}. However, further examination reveals differences in model performance. Interestingly, SBR appears to outperform the other models at certain time point predictions toward the middle and end of the study period ($\tilde{t} \approx 30,40$). This suggests that SBR may perform better were the smoothing method to have \textbf{not} been used. However, practically speaking, there is no circumstance in which one would want to discontinue use of such a method, but does seem to suggest that, intuitively, optimization methodology applying greater weight/emphasis to accurately learning ``earlier'' output nodes over ``later'' nodes may be beneficial.

\subsubsection{Visualizing Geographic Cluster Assignment}

Next, we briefly discuss the results of visualizing cluster assignment for $k=10,20,30,40$. These results can be observed in Figure \ref{fig:sc_map}, where each color represents a single cluster.

We first note that as $k$ increases, the elicited geographic regions become more precise, yet maintain geographic continuity. However, we secondly observe that some ZCTAs are not adjacent to any other ZCTA having the same cluster assignment. This disjointedness stems from the use of an adjacency representation of the affinity matrix on which spectral clustering is performed and is not unexpected. As $k$ increases it appears that the number of disjointed ZCTAs also increases. However, we see that the number of continuous regions also increases. In other words, while disjointedness seems to increase with $k$, the desired result of more localized continuous geographical regions is still achieved. Interestingly, when $k=40$, larger Iowa cities such as Des Moines (central Iowa) and Iowa City (central-eastern Iowa) begin to emerge.

\section{Related Work}

The topics related to and discussed throughout this work can best be categorized as \textit{disease and survival curve prediction} and \textit{geographic-based predictions and representation}.

There are many past works involving the prediction of diseases. These can be viewed as classification-based \cite{khosravi2015five,belciug2010two,ojha2017study,sandhuartificial,gupta2011data,belciug2013hybrid,puddu2012artificial} and survival-based \cite{cox1992regression,sharmasurvey,chi2007application, gupta2011data,katzman2016deep,samundeeswari2016artificial}. The focus of this work was on survival curve predictions. Such works can be examined by method, which include Cox proportional hazards model (CPH) \cite{cox1992regression}, which has been historically used to make such predictions, decision trees \cite{sharmasurvey}, and neural network-based models \cite{chi2007application, gupta2011data,katzman2016deep,samundeeswari2016artificial}, which are a more recent development. However, as Laurentiis and Ravdin \cite{de1994technique} point out, CPH has several caveats as compared to neural network-based approaches, including the naivety of the proportional hazards assumption and inability to capture nonlinear feature interactions. Furthermore, decision trees are constructed using greedy methodology and do not have the architectural benefits of neural networks. Hence, this work employed neural networks.

There are also many works focusing on \textit{geographic-based prediction and representation}. These works focus on incorporating geographical features into the predictive process. One method of representing geography is by fine grain lattice (i.e.,~grid) \cite{khezerlou2017traffic,lash2017large}. Such methods are akin to our SBR representation and suffer from the same shortcomings. Spatially adaptive filters \cite{tiwari2005using}, which can tie a single feature to geography when creating $\mathcal{M}$, which may be beneficial when the selected feature is particularly indicative of survival. This method would, however, still produce a binary feature representation, having the accompanying shortcomings discussed when disclosing SBR. Spectral clustering has been used to cluster both social networks \cite{white2005spectral} and for representing geo-spatial features \cite{frias2014spectral,van2013community}, as in this work, and produces a rich (i.e.,~non-sparse) vector of features.

\section{Conclusions and Future Work}

In this work we explored the use of two different geographical feature representations -- a simple binary representation (SBR) and a rich representation based on spectral clustering (which we term spectral analysis and methodologically refer to as RR-SA) -- to predict colorectal cancer survival curves for patients in the state of Iowa. We show that (a) survival curves can be reasonably estimated, although predictive performance deviates near the five-year survival mark, (b) the use of geographical features generally lead to better predictions, and (c) RR-SA trained models outperform those trained using SBR. Future work will involve exploration of different geographical representations, particularly those learned in conjunction with $\mathtt{g}^*$. Additionally, continued exploration of domains and scenarios in which SBR and RR-SA geographic representations provide benefit should be undertaken.

\section{Acknowledgements}

The authors would like to thank the Iowa Cancer Registry, State Health Registry of Iowa, and the Iowa Department of Public Health for the data. The authors would also like to thank Gary Hulett and Jason Brubaker for their help in dataset construction and Prakash Nadkarni for his help with both data acquisition and the IRB process.

%

\bibliographystyle{IEEEtran}
\bibliography{CancerReference}

\end{document}